\documentclass[letterpaper, 10pt, conference]{IEEEtran}
\pdfoutput=1
\IEEEoverridecommandlockouts                              

\usepackage{verbatim}
\usepackage{float}
\usepackage{amsmath}
\usepackage{graphicx}
\usepackage{amssymb}
\usepackage{array}
\usepackage[draft]{fixme}
\usepackage[english]{babel}
\usepackage{verbatim}
\usepackage{color}
\usepackage{multicol}
\usepackage{algorithm}
\usepackage[noend]{algpseudocode}

\makeatletter
\renewenvironment{subarray}[2][c]{%
  \if#1c\vcenter\else\vbox\fi\bgroup
  \Let@ \restore@math@cr \default@tag
  \baselineskip\fontdimen10 \scriptfont\tw@
  \advance\baselineskip\fontdimen12 \scriptfont\tw@
  \lineskip\thr@@\fontdimen8 \scriptfont\thr@@
  \lineskiplimit\lineskip
  \ialign\bgroup\ifx c#2\hfil\fi
    $\m@th\scriptstyle##$\hfil\crcr
}{%
  \crcr\egroup\egroup
}
\makeatother

\def\R{\mathbb{R}}

\newcommand{\ceil}[1]{\left\lceil #1 \right\rceil}

\newcommand{\frm}[1]{\langle #1\rangle}

\DeclareMathOperator{\sat}{sat} 

\title{Cooperative UAVs Gas Monitoring using Distributed Consensus}

\author{\IEEEauthorblockN{Daniele Facinelli, Matteo Larcher, Davide Brunelli, Daniele
  Fontanelli}
\IEEEauthorblockA{
Department of Industrial Engineering (DII)\\
University of Trento\\
 Via Sommarive 9, Trento, Italy \\
Email: \{name.surname\}@unitn.it}

 }

\makeatletter
\def\BState{\State\hskip-\ALG@thistlm}
\makeatother

\begin{document}
\bstctlcite{IEEEexample:BSTcontrol}

\maketitle

\thispagestyle{empty}
\pagestyle{empty}

\begin{abstract}
  This paper addresses the problem of target detection and
  localisation in a limited area using multiple coordinated
  agents. The swarm of Unmanned Aerial Vehicles (UAVs) determines the
  position of the dispersion of stack effluents to a gas plume in a
  certain production area as fast as possible, that makes the problem
  challenging to model and solve, because of the time variability of
  the target.  
	Three different exploration algorithms are designed
  and compared. 
	Besides the exploration strategies, the paper reports a solution for quick
  convergence towards the actual stack position once detected by one
  member of the team.  Both the navigation and localisation algorithms
  are fully distributed and based on the consensus theory. Simulations
  on realistic case studies are reported.
\end{abstract}


\section{Introduction and related works}
\label{sec:Introduction}

Unmanned Aerial Vehicles (UAVs), RPA (Remotely Pilot Aircraft) 
are nowadays used for a large variety
of applications, such as structural
monitoring~\cite{yoon2018structural}, emergency scenarios like search
and rescue~\cite{SeminaraF17mesas}, 
target tracking and encircling~\cite{MorelliVF17mesas}, reaching close to
commercial solutions in some cases~\cite{AmazonPJ}, or environmental monitoring~\cite{Rossi2014,IEEESJ2014}.  


The goal of this work is to localise the emission source of the gas dispersion as fast as possible as proposed in~\cite{SAS2015}, using 
multiple autonomous UAVs, and specifically, quadrotors~\cite{GPSaccuracy77}, to speed up the task of \textsl{seek-and-find}.
Our solution consists of two steps: the first concerns 
the exploration strategies towards the gas source detection, 
the second focuses on the distributed localisation once 
the source has been detected. 


The use of coordinated multiple agents is of course not novel in the
literature. In particular, it has been shown that the
information exchange between the agents of a team reduces to time to
accomplish a mission, which can be theoretically demonstrated, for
example applying the consensus theory~\cite{ren2005survey}.  In this
framework, UAV formation control using a completely distributed
approach among the controlled robots is a field already investigated
in the literature~\cite{zhu2015model}.  

Once the stack has been detected, the UAVs have to estimate its
position.  The literature for target tracking is quite
rich~\cite{li2005survey, AndreettoPMPF18mdpi}, especially
when the target is not governed by a white noise but, instead, has a
strong correlation between the executed
manoeuvres~\cite{singer1970estimating}.  Assuming that the target is
not moving but it is standing in a fixed unknown location has also
been presented in the literature, for example
by~\cite{li2014cooperative, AndreettoPFM15} using distributed
consensus-based known algorithms. 

%

%

Gas source localisation is considered in~\cite{mamduh2018gas} using
the ``Grey Wolf Optimiser'', a recently developed algorithm inspired
by grey wolves (Canis lupus).  It consists of three-stage procedure:
tracking the prey, encircling the prey, and attacking the prey.  This
algorithm works with a minimum of $4$ agents and, increasing the agent
number beyond seven, caused the reduction in success rate (percentage
of success in gas source localisation, $72\%$ with $7$ agents) and
increase the time to completion, due to increased frequency of
irrecoverable collisions between robots.

In this paper, three exploration algorithms for swarms of UAVs
conceived for gas source localisation are proposed and are based on i)
{\em coordinated scanning}, ii) {\em Random walk} and iii) {\em
	Brownian motion}.  All share the same method: as long as one agent
senses the gas with a concentration greater than a certain threshold,
it becomes the master of the group and the source localisation phase
starts.
%
In this latter phase, the agents are controlled with a distributed
algorithm that dynamically allocates the master role and localises the
source.  The algorithms presented in this paper converge on the
maximum concentration point, which is a property shared with the
Particle Swarm Operation~\cite{ferri2007explorative} and the Ant
Colony Optimisation~\cite{zou2009swarm} approaches.  Nonetheless,
there are some differences with respect to the discussed literature,
the most relevant being the approach adopted to find the source. In
fact, in the Grey Wolf Optimiser, agents start in a zone where the
concentration of the gas is above a minimum sensing threshold, and
this data is necessary to reach the source; in our case, instead, UAVs
can start in any part of the environment.  Moreover, we can
statistically prove through extensive simulations that our algorithms
can operate also with 2 drones instead of a minimum of 4 and that the
rate of success under the presented assumptions is 100\%.
%
%
Finally, for the estimation problem as well as for the exploration, we
have considered a limited sensing range for the UAVs. 

The paper is organised as follows.  Section~\ref{sec:Models}
introduces the adopted models and presents the problem at hand. Then,
in Section~\ref{sec:Exploration} and in Section~\ref{sec:Localisation}
the algorithms used for the area exploration and the source
localisation are defined. Section~\ref{sec:Results} presents extensive
simulations analysing all the relevant features of the algorithms and
their comparison.  Finally, Section~\ref{sec:Conclusions} draws the
conclusions and discusses possible future developments.

\section{Problem Formulation and Adopted Models}
\label{sec:Models}

The problem we are aiming at is the detection and localisation of a
certain phenomenon (pollutant leak) taking place in an unknown
location (stack) inside the environment of interest.  The phenomenon
is supposed to be measurable by a sensor rigidly fixed on the robot
chassis and capturing the intensity of the pollutant concentration.
More precisely, if $\frm{G} = \{O_g,\, X_g,\, Y_g,\, Z_g\}$ is a fixed
{\em ground} reference frame, where $O_g$ identifies the origin of
$\frm{G}$, the drone can be modelled as an unconstrained rigid body
having $6$ degrees of freedom. The position of its centre of mass in
$\frm{G}$ (which will be considered as the reference point of the
drone) is given by $\xi = [x, y, z]^T$.
Considering that the UAV dynamics is given by a quadrotor and
recalling that quadrotors are controlled by differentially driving the
four motors, we will make use of the results in~\cite{DronesTheory} to
decoupling the attitude and position control.
In practice, given a desired position $\xi^*$ to be reached, it is
possible to generate a vertical thrust $u_1$ and a torque vector $u_2$
around its centre of mass to steer the UAV towards $\xi^*$.  To this
end, a suitable trajectory planner generating smooth trajectories
connecting the starting and ending position of the quadrotor is
utilised (this component is not described in this paper, but it is
quite customary in the literature).  The UAV model thus considered is
approximated, since:

\begin{itemize}

\item The model does not take into account actuation saturations,
  while in reality this is a major issue for this robotic systems.
  Therefore, we assume the linear velocity limited to $3$~m/s;

\item We have supposed to be capable of measuring the whole state. In
  practice a nonlinear observer is adopted~\cite{DronesTheory};

\item We have not modelled several aerodynamic
  effects~\cite{DronesTheory}. For instance, if the robot is flying at
  high speed, they become no more negligible.

\end{itemize}

We consider these approximations acceptable in this work since we are
more interested in the distributed coordination of the UAVs formation,
making them a team of autonomous agents, instead of giving a too much
detailed analysis of the single agent dynamic.  As a result, the
individual UAV can be considered to be controlled by generating a set
of desired configurations to be reached.  In particular, we will
assume that the $i$--th agent dynamic, with $i = 1,\dots, n$ is given
as a first order integrator, i.e.
\begin{equation}
  \label{eq:LinearDyn}
  \dot \xi_i(t) = u_i(t) ,
\end{equation}
where $u_i\in\R^3$ are the three independent control inputs, hence
$\xi_i = [x_i, y_i, z_i]^T$. The UAVs are supposed to be equipped with
GPS sensors to determine the UAV location with negligible uncertainty.
For the agents orientation, a magnetic compass is considered.  As a
consequence, we will refer to $\xi_i$ as the actual position of the
$i$-th quadrotor in the reference frame $\frm{G}$.

\subsection{Gaussian Plume}
\label{subsec:GasModel}

\begin{figure}[t]
  \begin{centering}
    \includegraphics[width=0.6\columnwidth]{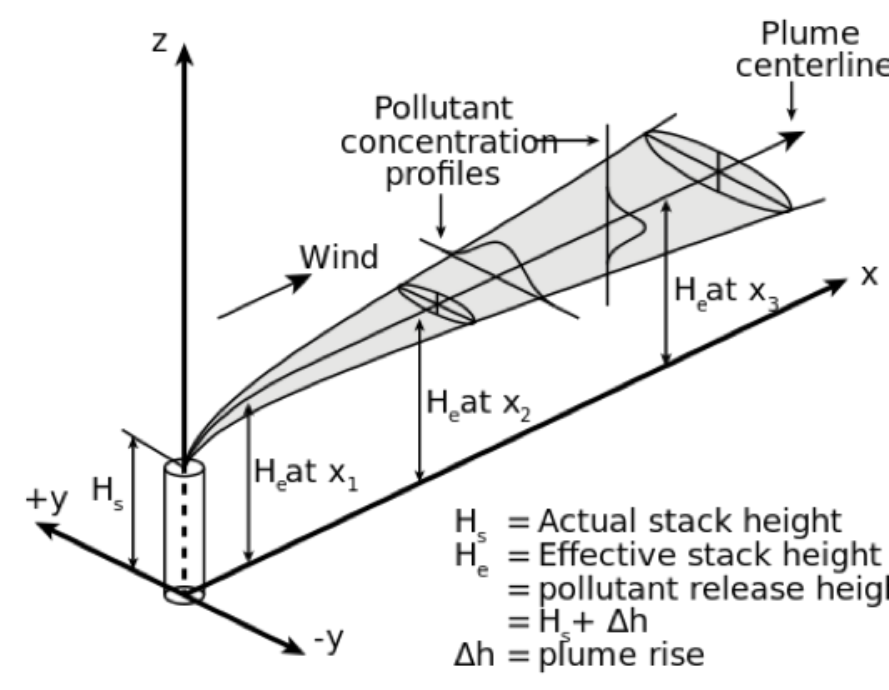}
    \caption{Visualisation of a buoyant Gaussian air pollutant
      dispersion plume~\cite{GaussianPlume}.}
    \label{fig:GaussianPlume}
  \end{centering}
	\vspace{-6mm}
\end{figure}

Gaussian Plume~\cite{GaussianPlume} is a well-known method used to
simulate the behaviour of pollutant mixture released in atmosphere by a
source in position $p_s = [0,0,H_s]^T$ expressed in $\frm{G}$, under
the effect of the wind (see \figurename~\ref{fig:GaussianPlume}).
With the wind blowing, e.g., in the $X_g$ direction with speed $w$
measured in meters per second, the plume spreads as it moves along
$X_g$.  In particular, the local concentration $C(x,y,z)$ at a given
point $(x,y,z)$ forms distributions which have shapes that resembles
Gaussian distributions along the $Y_g$ and $Z_g$ axes as described in
\figurename~\ref{fig:GaussianPlume}.  
The two modelling distributions are given respectively by
$p_\star(\star) = \frac{1}{\sqrt{2\pi}\sigma_\star(x)}
e^{-\frac{\star^2}{2\sigma_\star^2(x)}}$, with $\star = y,z$ and
where the standard deviations $\sigma_y(x)$ and $\sigma_z(x)$ of these
Gaussian probability density functions (pdfs) 
are a function of $x$, i.e. the spread increases with the distance $x$
from the source.  At atmospheric stability conditions and using 
the so called Pasquill-Gifford Curves~\cite{davidson1990modified}, we
have $\sigma_y(x) = c x^d$ and $\sigma_z(x) = a x^b$, whose parameters
can be found in the well-known Pasquill Gifford stability
tables~\cite{davidson1990modified}.

Assuming as customary the two random variables modelling the plume
spread independent, we have that the joint pdf
$p_{y,z}(y,z) = p_y(y)p_z(z)$ and hence the concentration is
\begin{equation}
\label{eq:Conc}
\begin{aligned}
C(x,y,z,H_s) = & \frac{Q}{w} \frac{1}{2\pi \sigma_y(x) \sigma_z(x)}
e^{-\frac{y^2}{2\sigma_y^2(x)}} \cdot \\
& \cdot \left [
e^{-\frac{(z - H_s)^2}{2\sigma_z^2(x)}} + e^{-\frac{(z +
		H_s)^2}{2\sigma_z^2(x)}} \right ] ,
\end{aligned}
\end{equation}
which accounts for the ground effect and where $Q$ is the dispersion
mass and $w$ is the wind speed.


For the concentration sensor, the MiniPID 2
PID\footnote{https://www.ionscience.com/products/minipid-2-hs-pid-sensor/minipid-2-hs-pid-sensor-whats-included/}
sensor, by Ion Science, is supposed to be available. This type of
device is a PID (photoionisation detector) sensor, able to sense
volatile organic compounds in challenging environments.
Assuming that a gas source is positioned in $p_s = [x_s, y_s, H_s]^T$
and that the $i$--th robot is located in position $\xi_i$, the gas
concentration measured in the sensor position (here assumed co-located
with the UAV centre of mass without loss of generality) is a modified
version of~\eqref{eq:Conc}, i.e.
\begin{equation}
  \label{eq:ConcFinal}
  \begin{aligned}
    C_i(\xi_i,p_s) = & \frac{Q}{w} \frac{1}{2\pi \sigma_y(x_i - x_s)
      \sigma_z(x_i - x_s)}
    e^{-\frac{(y_i - y_s)^2}{2\sigma_y^2(x_i - x_s)}} \cdot \\
    & \cdot \left [ e^{-\frac{(z_i - H_s)^2}{2\sigma_z^2(x_i - x_s)}} +
      e^{-\frac{(z_i + H_s)^2}{2\sigma_z^2(x_i - x_s)}} \right ] .
  \end{aligned}
\end{equation}
Considering the limited range of the sensor, we finally have that the
measurement results are given by
\begin{equation}
  \label{eq:ConcFinal2}
  h_i(\xi_i) = \mbox{sat}\left (C_i(\xi_i,p_s)\right) ,
\end{equation}
where $\sat(\alpha)$ is equal to $\alpha$ if $\alpha < 100$, otherwise
$\alpha = 100$ (maximum sensing range).

\subsection{Problem Formulation}
\label{subsec:ProblemFormulation}

Consider a set of $n$ UAVs flighting at the same height, i.e. on a
generic $X_g \times Y_g$ plane, and a gas source position
$p_s = [x_s, y_s, H_s]^T$.  Assuming that the $i$--th robot is located
in position $\xi_i$, the problem is to estimate the plane position
$\hat p_s = [\hat x_s, \hat y_s]^T$ of the source with an error
$e_s = [x_s - \hat x_s, y_s - \hat y_s]^T$ such that
$\|e_s\|\leq \rho_m$.

\section{Exploration Strategy for Stack Detection}
\label{sec:Exploration}

The concept of multiple autonomous systems, i.e. computerised systems
composed of multiple intelligent agents that can interact with each
other through exchanging information, has been proposed for instance
in~\cite{AndreettoPMPF18mdpi}. The use of multiple autonomous systems
is usually more efficient than single agent systems from both the task
performance and the computational efficiency viewpoints, besides the
fact that some problems are difficult or impossible for single
agents. Regarding coordination between agents, consensus control
approaches have gained a lot of attention for autonomous agents
coordination~\cite{ren2005survey}.  The main objective is to find a
consensus algorithm between the multiple autonomous agents making use
of local information and with the aim to control the group using
agreements schemes.  

Assuming the simple linear dynamic in~\eqref{eq:LinearDyn}, a quite
simple consensus algorithm that let the $n$ agents to converge in a
fixed position (rendezvous problem) is the following:
\[
  \overline{u}_i(t) = \sum_{j=1}^n l_{ij}(\xi_i(t) - \xi_j(t)) ,
\]
where $l_{ij}$ is the element in position $(i,j)$ of the Laplacian
matrix; 
if the final positions have
to be separated by relative position vectors $\delta_i$, then
\begin{equation}
  \label{eq:SimpleCons}
  \overline{u}_i(t) = \sum_{j=1}^n l_{ij}\left [(\xi_i(t) - \xi_j(t)) -
    (\delta_i- \delta_j)\right ] ,
\end{equation}
i.e., the UAVs reaches a final fixed formation. From this simple idea,
many different approaches can be applied using the concepts of
adjacency and Laplacian matrices~\cite{ren2005survey}.

\subsection{Coordinated Scanning}

One simple and effective approach is to use a leader-follower
formation: the leader is in charge of moving inside the environment
using a controlled reactive behaviour or just following a predefined
trajectory (that is the solution here considered), while the followers
are adapt to the leader's motion to preserve the formation.
Using~\cite{huang2017consensus} and assuming that the $i$--th agent is
the leader, it is possible to use
\begin{equation}
  \label{eq:LeaderEq}
  u_i(t) = \dot\xi_d(t) - k_i (\xi_i(t) - \xi_d(t) - \delta_i) +
  \overline{u}_i(t) ,
\end{equation}
where $\overline{u}_i(t)$ is reported in~\eqref{eq:SimpleCons},
$\xi_d(t)$ is the desired trajectory, $\dot \xi_d(t)$ the desired
trajectory dynamic and $k_i > 0$ is a tuning parameter.  Similarly,
for the $k$-th follower, with $k \neq i$,
\begin{equation}
  \label{eq:FollowerEq}
  u_k(t) = \frac{1}{\sum_{j=1}^n a_{ij}} \left [ \sum_{j=1}^n a_{ij}
    \dot \xi_j(t) +
  \overline{u}_i(t) \right ] ,
\end{equation}
where, again, $\overline{u}_k(t)$ is in~\eqref{eq:SimpleCons} and
$a_{ij}$ is the element in position $(i,j)$ of the adjacency
matrix. 
Notice that this algorithm is entirely distributed and the information
that the $k$-th agent shares are its position $\xi_i(t)$, velocity
$\dot\xi_i(t)$, and its formation vector $\delta_i$.

\begin{figure}[t]
  \begin{centering}
    \includegraphics[width=0.8\columnwidth]{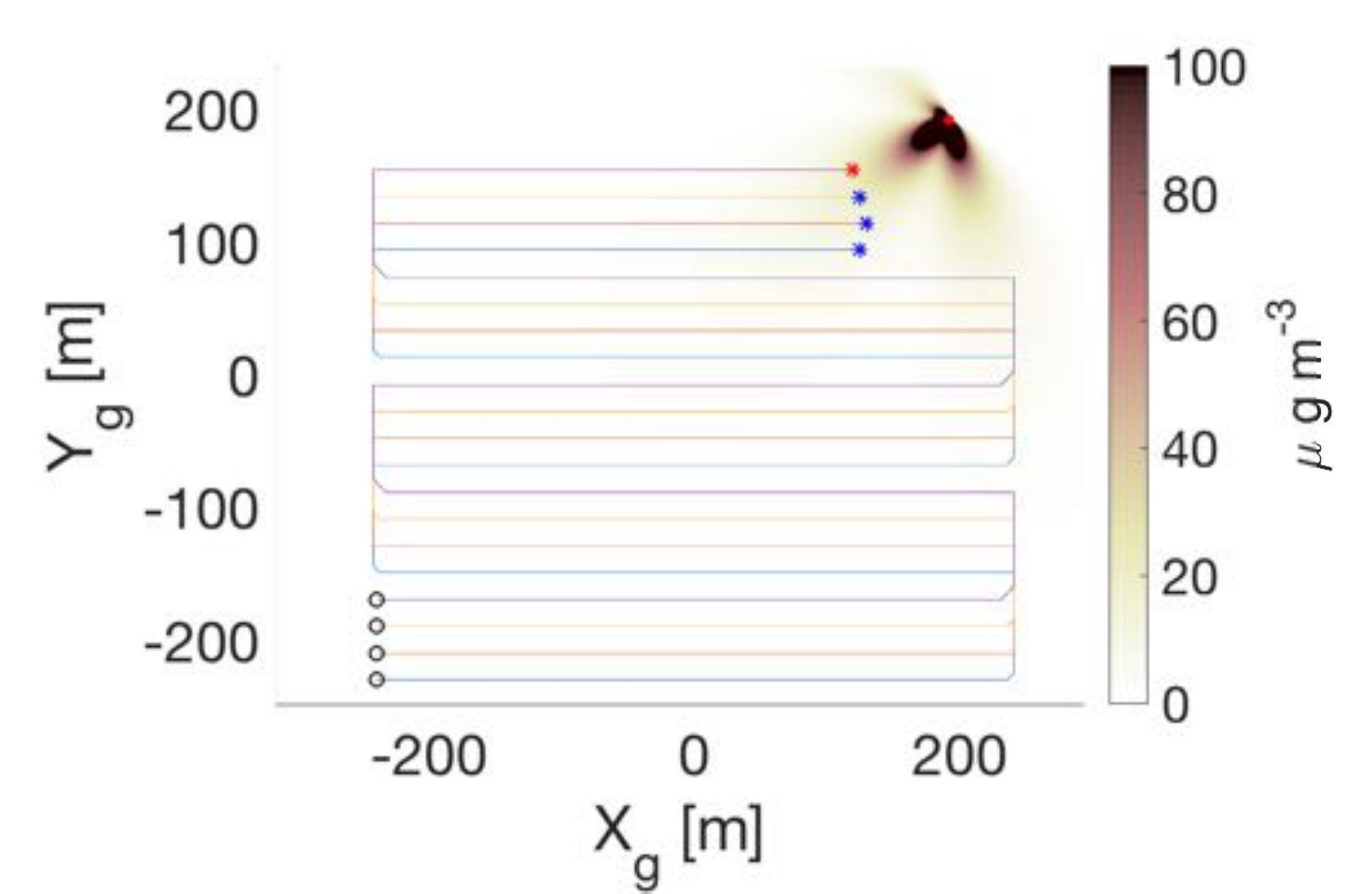}
    \caption{Example of coordinated scanning. UAVs start from an
      initial configuration (circles) that is a line and scan the
      whole area until one of the team members (red star) does not
      detect a sufficiently high gas concentration (stars in the final
      configuration).}
    \label{fig:Scanning}
  \end{centering}
\end{figure}

For the coordinated scanning, all the agents can start from an
arbitrary position and using~\eqref{eq:SimpleCons} converge to a
line. In particular, we select as leader the $l = \ceil{(n/2)}$ agent
(any other heuristic for the leader choice can be applied), which is
in fixed position, i.e. $u_l(t) = 0$, and has $\delta_l = [0,0,0]^T$.
For any other agent $i \neq l$, we make use of~\eqref{eq:SimpleCons}
with $\delta_i = [0, -(l - i) \Delta_y, 0]^T$, 
where $\Delta_y = 20$~m is the desired distance along the $Y_g$ axis.
This way, the UAVs are placed on a line parallel to $Y_g$ (see
\figurename~\ref{fig:Scanning}).
Once this initial configuration is reached, the exploration starts.
In the example of \figurename~\ref{fig:Scanning}, the exploration
follows lines that are parallel to the $X_g$ axis, hence the leader
desired velocity in~\eqref{eq:LeaderEq} is set to
$\dot\xi_d = [\pm v_m, 0, 0]^T$ for motions parallel to the $X_g$
direction or to $\dot\xi_d = [0, v_m, 0]^T$ when are moving upwards
along the $Y_g$ direction.
In the simulations of \figurename~\ref{fig:Scanning} the UAV
velocities are set to $v_m = 3$~m/s and a wedge formation is chosen.
The decision to switch between the different directions is taken by
the leader knowing the dimension of the area to scan, while the motion
switches from vertical to horizontal whenever the leader has travelled
for a distance of $d = n \Delta_y$.

\subsection{Random Walk}

\begin{figure}[t]
  \begin{centering}
    \includegraphics[width=0.8\columnwidth]{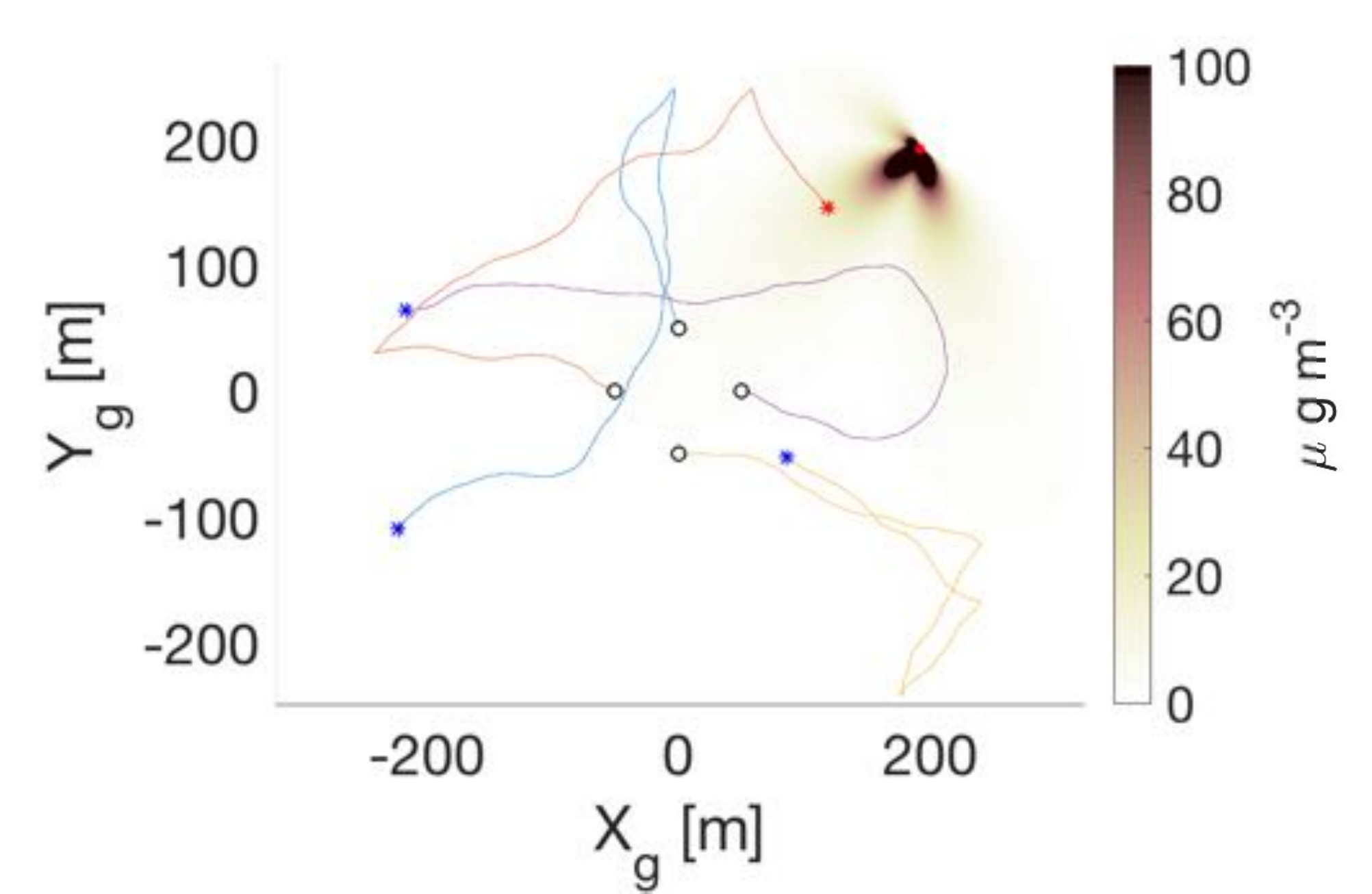}
    \caption{Example of random walk algorithm. UAVs start from an
      initial configuration (circles) that is a circle centred at the
      origin and scan the area until one of the team members (red
      star) does not detect a sufficiently high gas concentration
      (stars in the final configuration).}
    \label{fig:RandomWalkAndBrownian}
  \end{centering}
\end{figure}

We simulate a random walk for the motion of the UAVs inside the area
of interest.  In this case, the $i$--th agent moves according to this
dynamic
\[
  \dot\xi_i(t) = \begin{bmatrix}
    \cos(\theta_i(t)) \\
    \sin(\theta_i(t)) \\
    0
    \end{bmatrix} v_m ,
\]
that is computed every $\Delta_t$ seconds and in which the orientation
$\theta_i(t)$ is updated following the update rule $\theta_i(t) =
\theta_i(t - \Delta_t) + \nu$, 
being $\nu\sim\mathcal{U}(-\theta_M,\theta_M)$ a random variable
uniformly distributed and generated by a white stochastic process,
where $\theta_M$ is a user defined constant. Hence, the name random
walk for this exploration strategy.  Whenever the $i$-th agent reaches
the area of interest border, a new orientation is randomly generated
according to $ \theta_i(t) = \nu \theta_{\mbox{in}}
\theta_r/\theta_M$, 
where $\theta_{\mbox{in}}$ is the orientation of the angle of the
local normal vector of the border and pointing inwards the region,
while $\theta_r$ is a user defined constant threshold describing the
feasible orientations originating from a border location.  For this
particular strategy, a collision avoidance mechanism is needed (recall
that all the UAVs are moving at the same height).  In this paper, we
adopt a quite straightforward solution: whenever the distance between
the $i$--th and the $j$--th agent is less than a given safety
threshold $d_m$, i.e., $\| \xi_i(t) - \xi_j(t)\|_2 \leq d_m$, the
agents are subjected to a repulsive force along the directions
$\theta_i(t) = \arctan\left(\frac{y_i(t) - y_j(t)}{x_i(t) -
    x_j(t)}\right )$ and $\theta_j(t) = \theta_i(t) + \pi$ .
A simulation explaining this behaviour is reported in
\figurename~\ref{fig:RandomWalkAndBrownian}-(a), where
$\theta_M = 10^\circ$ and $\theta_r = 75^\circ$.

The initial configuration (circles) is on a circle centred at the
origin and with radius $50$~m.

\subsection{Brownian Motion}

The final exploration strategy is dubbed Brownian motion.  According
to the basic rules of Brownian motion particles, in this work the
random change in direction of the agents is determined by two events:
collision avoidance intervention and boundary collision.

\section{Localisation of the Stack}
\label{sec:Localisation}

Since all the UAVs in the team has the capability of sensing the gas
concentration using the sensor described in
Section~\ref{subsec:GasModel}, once any of the team members detects it,
it becomes the new leader of the formation (signed with a red star in
\figurename~\ref{fig:Scanning} and
\figurename~\ref{fig:RandomWalkAndBrownian}). This type of
information, together with position data, are shared with other agents
and the localisation phase starts.  The algorithm works as follows:

\begin{enumerate}

\item At the beginning, the team is arranged in a circle of radius $r$
  with the current leader acting as the fixed circle centre.  The
  positions on the circle are determined by letting the leader, say
  the $j$--th agent, computing the position vectors $\delta_i$,
  $i \neq j$ using polar coordinates, for all the agents and then
  applying~\eqref{eq:FollowerEq};

\item When the circle formation is reached, the $n-1$ UAVs move along
  a logarithmic spiral with the $i$-th agent radius
  $r_i = r e^{b \beta_i}$,
  where $\beta_i$ is the arc spanned by the $i$-th agent around the
  leader from the initial circle position and the fixed coefficient
  $b$ is computed in order to have a reduction on the radius between
  two agents position of $\gamma \rho_m$, where $\rho_m$ is the
  maximum allowed error defined in
  Section~\ref{subsec:ProblemFormulation}.  In particular,
  $b = \frac{n-1}{2\pi}\log\frac{r - \gamma \rho_m}{r}$;

\item If the $i$--th agent along the spiral motion detects a gas
  concentration larger than the actual leader, i.e.
  \[
    h_i(\xi_i) > h_j(\xi_j) ,
  \]
  then the $i$--th agent becomes the leader and the algorithm starts
  over from Step~1;

\item The spiral motion for all the agents ends when $r_i = \rho_m$,
  i.e. the desired tolerance of
  Section~\ref{subsec:ProblemFormulation};

\item At this point, the agents start to move on the circle with
  radius $\rho_m$ for an arc of length $2\pi/(n-1)$;

\item If the $i$--th agent along this arc-circle motion verifies the
  gas concentration condition $h_i(\xi_i) > h_j(\xi_j)$, the $i$--th
  agent becomes the leader and the algorithm restarts from
  Step~5. Otherwise the algorithm ends.

\end{enumerate}

The algorithm thus designed works assuming that the gas concentration
is a decreasing function of the distance from the source, which is
verified by the Gaussian plume model described in
Section~\ref{subsec:GasModel}.

\section{Simulation Results}
\label{sec:Results}


\begin{figure*}[ht]
  \begin{centering}
    \begin{tabular}{cc}
      \includegraphics[width=0.8\columnwidth]{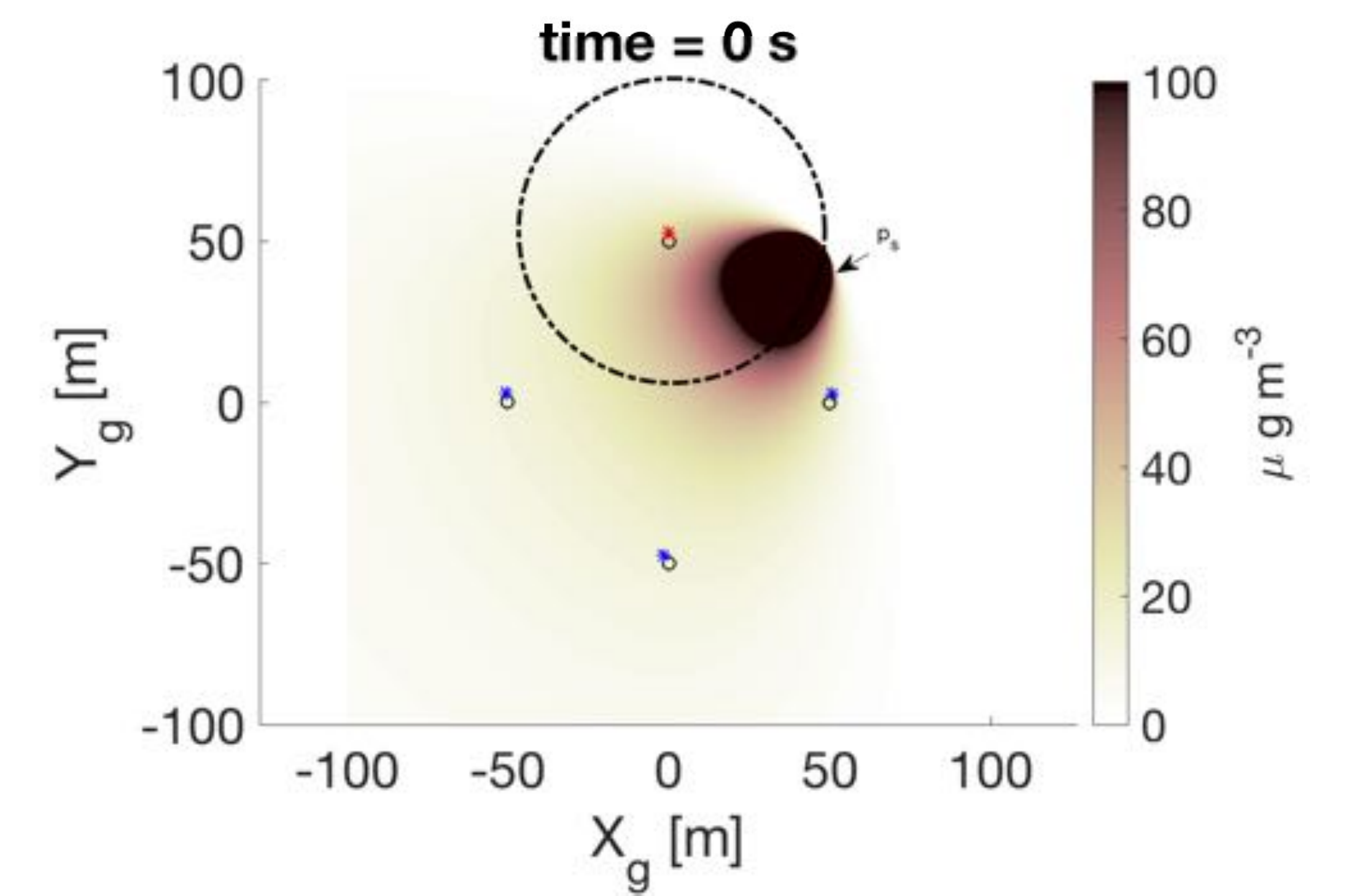}
      &
        \includegraphics[width=0.8\columnwidth]{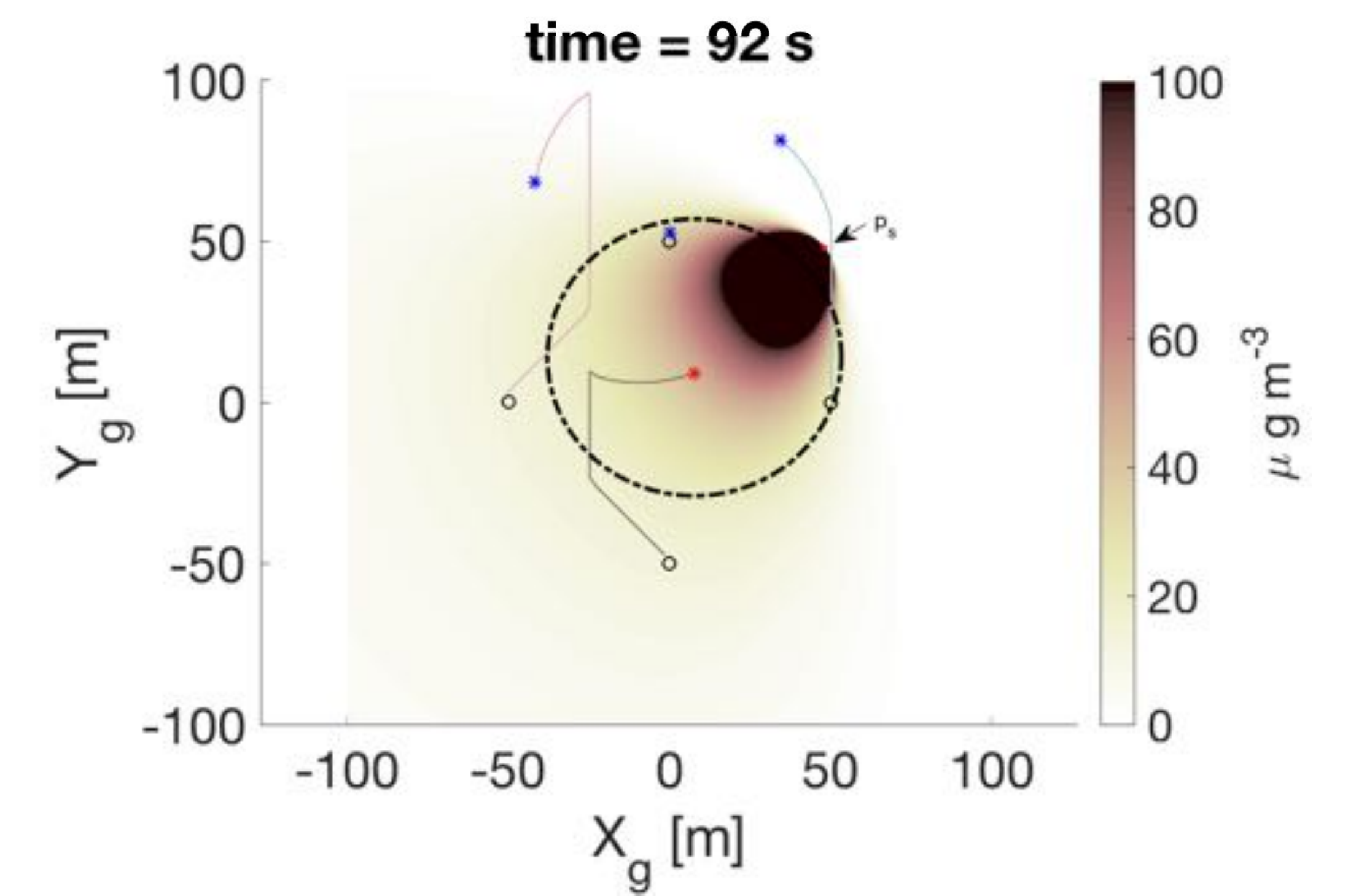}
      \\
      (a) & (b) \\
      \includegraphics[width=0.8\columnwidth]{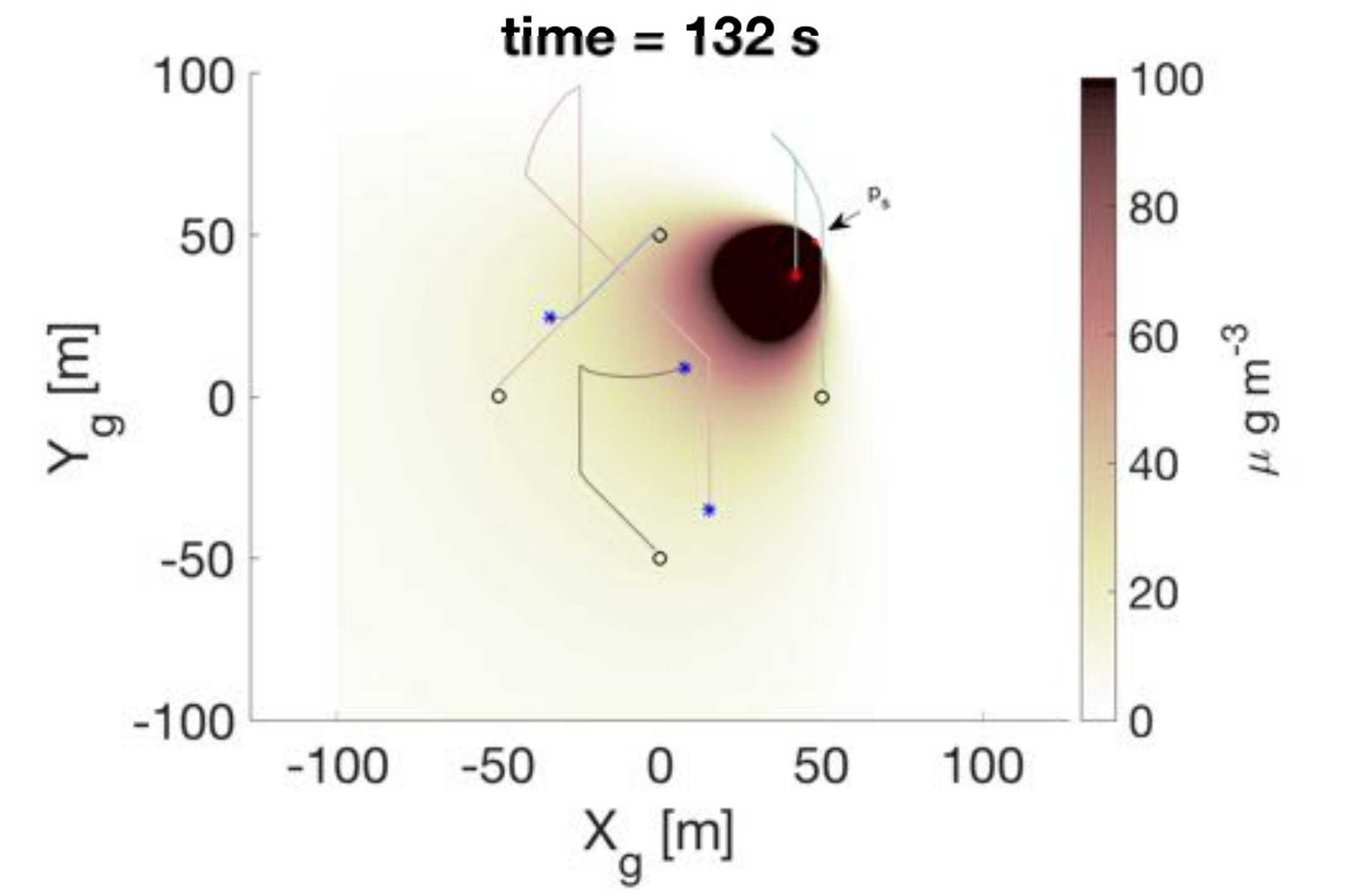}
      &
        \includegraphics[width=0.8\columnwidth]{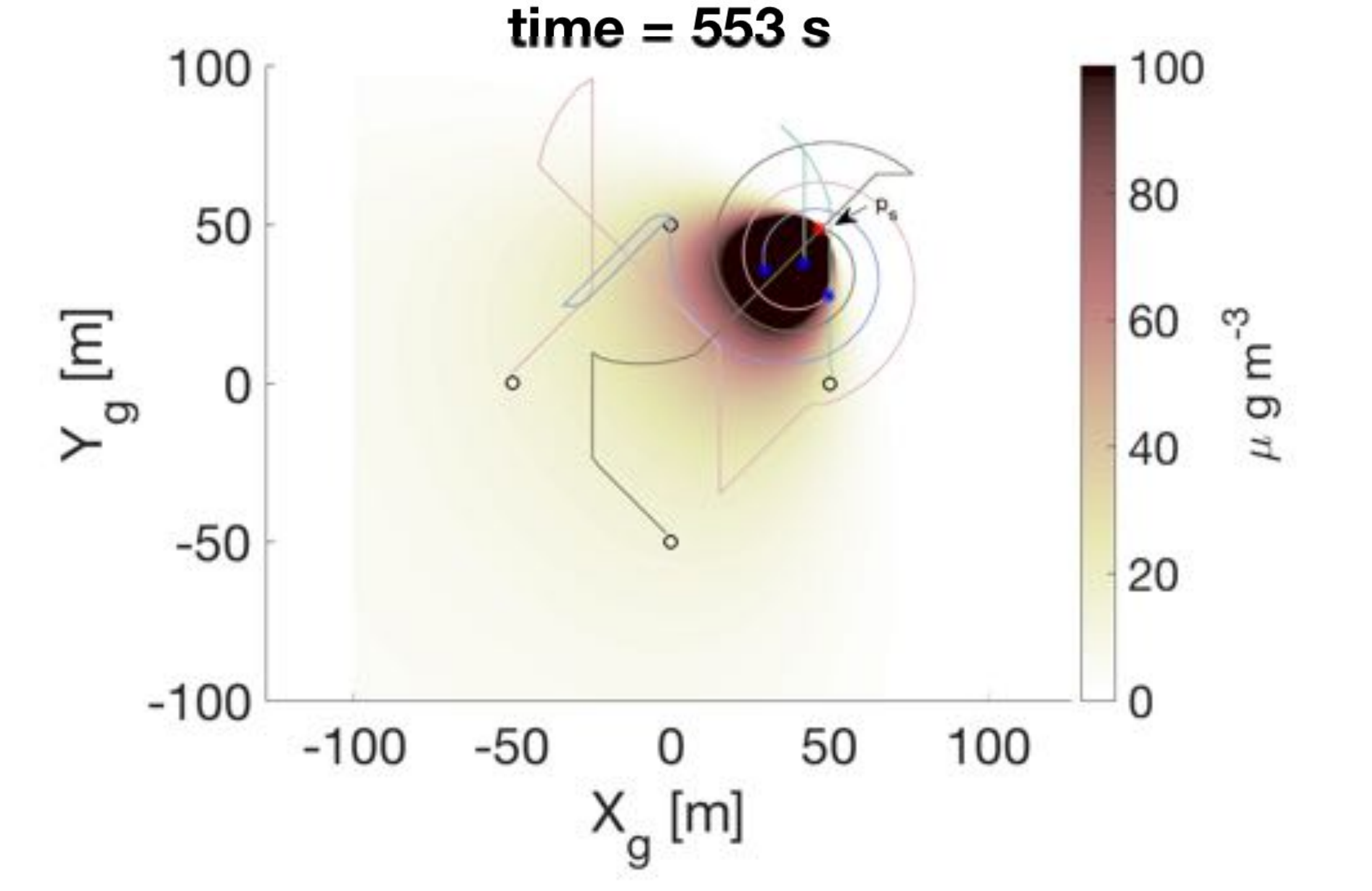}
      \\
      (c) & (d) \\
    \end{tabular}
    \caption{Four UAVs trajectories during the localisation phase of
      the stack in position $p_s$.  (a) At the beginning, the UAVs are
      requested to move on a circle (dash-dotted line) around the
      leader (red star). (b) After 92 seconds, the leader changes (red
      star), with a new desired circle (dash-dotted line). (c) After
      $132$~seconds, a new leader reaches the desired circle and
      immediately becomes the new leader (red star). (d) After moving
      on logarithmic spirals, a new leader is detected, which is quite
      close to the source $p_s$.  The overall trajectories followed by
      the four UAVs are depicted with solid lines.}
    \label{fig:Localisation}
  \end{centering}
\end{figure*}

In this section, we propose a statistical comparison and analysis of
the effectiveness of the designed algorithms.  This simulative
analysis is strictly necessary since many variables come into play,
i.e. the different exploration strategies, the behaviour of the
Gaussian plume, the number of UAVs.  In all the simulation here
presented we assume a maximum UAV velocity of $v_m = 3$~m/s, a stack
height $H_s = 3$~m (which is equal to the agents flighting height), a
stack emission of $5$~g/s and a searching area of
about 
$880\times 880$~m$^2$. 
Statistical data have been collected along $m_s = 100$ simulations in
each considered configuration, $m_w = 100$ different Gaussian plume
conditions as well as a variable number of UAVs, from $2$ to $20$
agents.  Depending on the particular simulation set-up, each
simulation lasts for a variable number of $m_t$ samples, where the
sampling time is denoted as $\Delta_t = 1$~s as in
Section~\ref{sec:Exploration}.

\subsection{Pasquill-Gifford model}

First, we present the results of the Pasquill-Gifford model presented
in Section~\ref{subsec:GasModel}.  The gas dispersion has been
modelled considering different conditions. Inspired
by~\cite{Connolly11}, a synthetic dataset has been generated based on
experimental observations: a) wind coming from a constant direction;
b) wind coming from a completely random direction; c) wind coming from
a prevailing random
direction. 
All the simulations have been carried out considering very unstable
wind conditions.  The concentration of the gas mixture in every point
of the area are time averaged along the simulation in order to avoid
transient conditions happening at the very beginning of the gas leak
emission.  \figurename~\ref{fig:Localisation} depicts the simulation
results
with colours identifying the concentration and the red dot
representing the position of the source.

\subsection{Exploration Strategies}

\begin{figure}[t]
  \begin{centering}
      \includegraphics[width=\columnwidth]{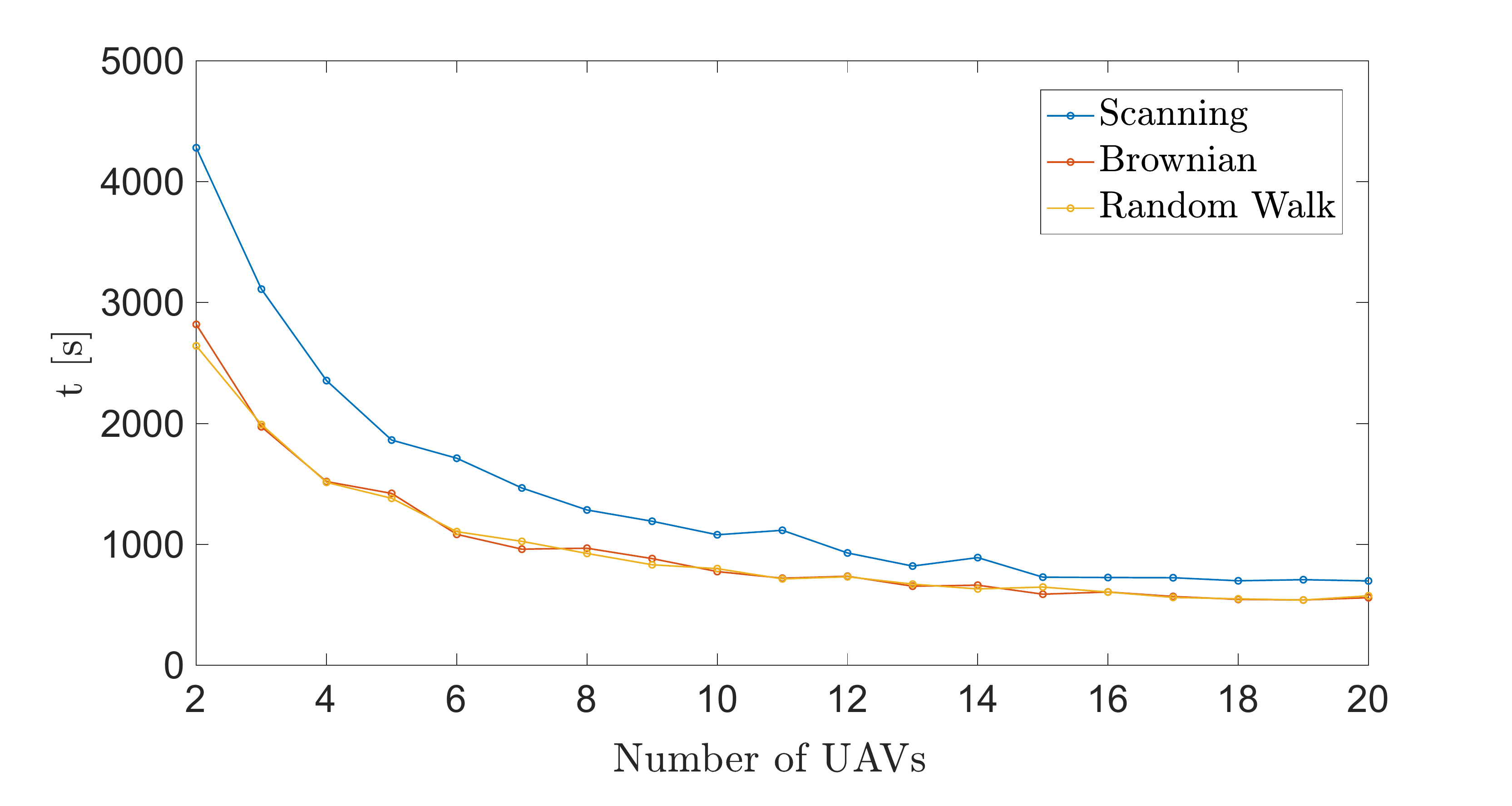}
      \caption{Time to accomplish the mission as a function of the
        number of UAVs.}
    \label{fig:TimeExploration}
  \end{centering}
\vspace{-4 mm}
\end{figure}

For the statistical analysis of the three different exploration
strategies presented in Section~\ref{sec:Exploration} the minimum
detection threshold for each agent is $h_m = 30$ parts per billion,
i.e. conservatively $6$ times the rated sensitivity of the sensor
described in Section~\ref{subsec:GasModel}.
\figurename~\ref{fig:TimeExploration} depicts the mean average taken
to complete the mission, i.e. until $\exists i$ such that
$h_i(\xi_i) > h_m$ as a function of the number of UAVs.

Trivially, all the different algorithms behave better increasing the
number of drones. Another important aspect that is underlined by the
graph is the fact that an algorithm based a deterministic approach has
typically worse performance than random exploration approaches, such
as Brownian motion and random walk.

\subsection{Localisation of the Stack}

We will now present the results of the source localisation algorithm.
The maximum tolerable error is set to $\rho_m = 5$~m.  The radius of
the circle arranged at the beginning of the localisation phase is set
to $r = 50$~m.  The ratio governing the contraction of the radius for
the logarithmic spirals is set to $\gamma = 1/4$.  An example of a
localisation manoeuvre for $4$ UAVs is reported in
\figurename~\ref{fig:Localisation}.  At the beginning
(\figurename~\ref{fig:Localisation}-(a)), the UAVs are requested to
move on a circle (dash-dotted line) around the leader (red star),
which is the first sensing a concentration greater than $h_m$. After
that all the UAVs have reached the circle and started to move along
the spiral (after $92$~s, \figurename~\ref{fig:Localisation}-(b)), the
leader changes (red star) and a new desired circle (dash-dotted line)
is established. The UAVs start moving toward the new circle and, right
after $40$~second,s they reached it: a new leader is then immediately
determined (red star in \figurename~\ref{fig:Localisation}-(c)). At
this point, the UAVs move toward the new circle and start the spiral
motions.  Then, at time $553$~s, one of the UAVs senses a high level
of concentration, and becomes the new leader, denoted with the red
star in \figurename~\ref{fig:Localisation}-(d).  Even though this new
leader is quite close to the source $p_s$, the process continues in
this way, changing two additional leaders and ending after $721$~s.

\begin{figure}[t]
	\begin{centering}
		\includegraphics[width=\columnwidth]{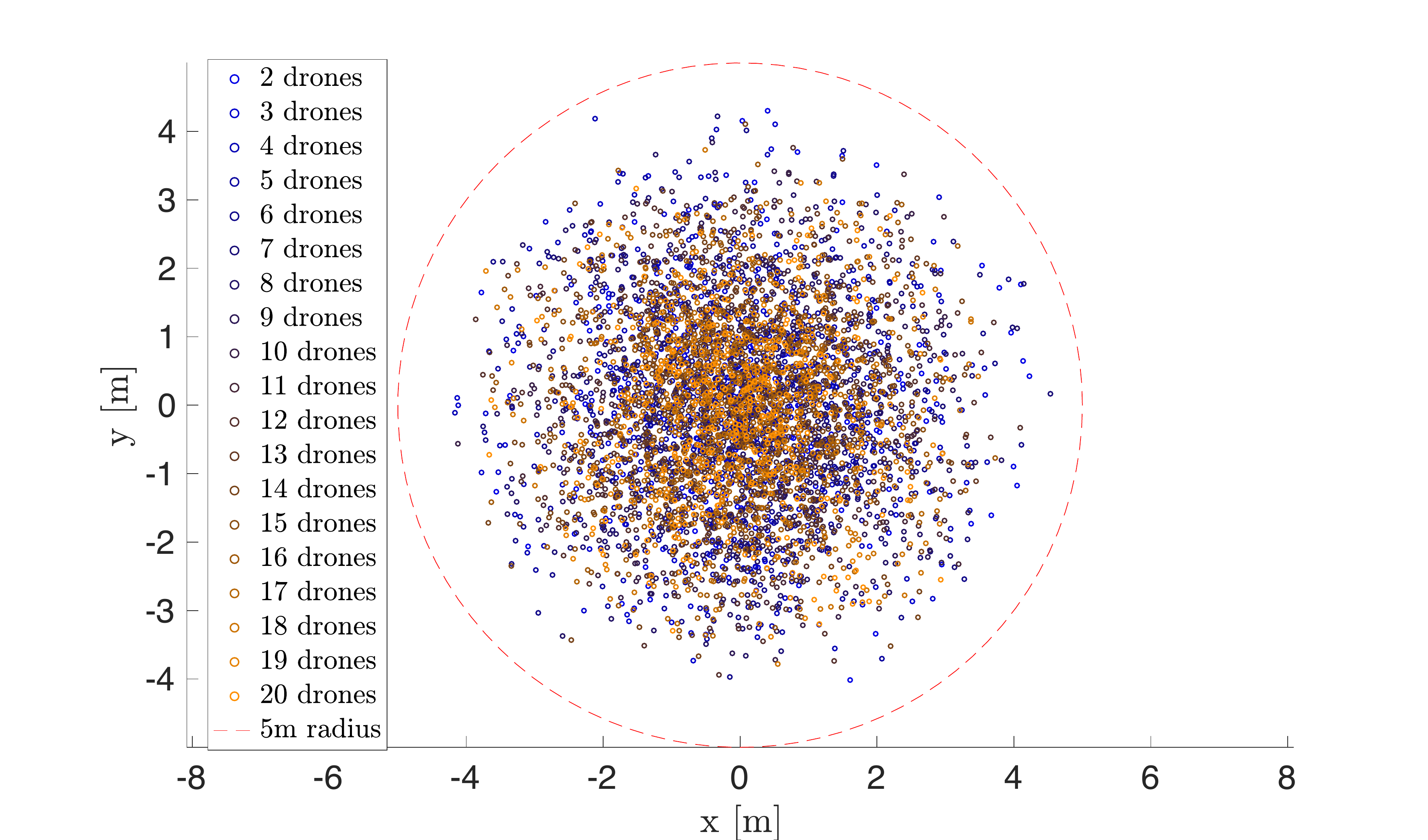} \\
		\caption{Scattering
			plot of the error $\|e_s\|$.}
		\label{fig:DistError}
	\end{centering}
\vspace{-3mm}
\end{figure}

To clearly state the performance of this approach, we will start by
first saying that the average of
the 
localisation error $\|e_s\|$ defined in
Section~\ref{subsec:ProblemFormulation} along all the simulations
does not vary in a significant way for the three algorithms and it
stabilises around $1.6$~m $< \rho_m$.  Moreover,
\figurename~\ref{fig:DistError} shows that the condition
$\|e_s\| < \rho_m$ is always verified by design.
As a concluding remark, the exploration strategy adopted does not play
any role in the final accuracy, while the number of UAVs reduces the
exploration time.

\section{Conclusions}
\label{sec:Conclusions}

In this paper we have designed and analysed a multi-agent system
conceived for the detection of a source of a gas dispersion. Methods
based on random motions, such as Brownian motion and random walk here
presented, perform averagely better in terms of completion time with
respect to deterministic approaches, such as the coordinated scanning
here proposed.  For the maximum time of mission completion,
deterministic approaches offer stringent guarantees, while stochastic
approaches can be sometimes excessively high.  The final accuracy of
the source position is based on a common distributed algorithm
designed on purpose.  Moreover, we have shown how the number of agents
in the team improves the exploration algorithms performance, while it
has negligible effects on the localisation accuracy.

Future research directions will be focused on the actual deployment of
the algorithms in real scenarios, on the definition of more flexible
distributed control approaches as well as on the multiple sources
scenarios.

\bibliographystyle{IEEEtran}
\bibliography{reference}

\end{document}